\documentclass{article}
\usepackage[T1]{fontenc}
\usepackage{spconf,amsmath,amssymb,graphicx,hyperref}
\usepackage{booktabs}
\usepackage{multirow} 
\usepackage{graphicx}
\usepackage[table]{xcolor}
\usepackage{booktabs}
\usepackage{multirow}
\usepackage{makecell}
\usepackage[table]{xcolor}
\title{AgriCountDINO: Parameter-Efficient Exemplar-Guided Counting and Localization in Agriculture}
\name{
Shengjie Guo$^{1,3}$,
Xin Li$^{2}$,
Borjana Arsova$^{3}$,
Hanno Scharr$^{4}$,
Silvio Salvi$^{1}$\sthanks{Corresponding Author}
}

\address{
\scalebox{0.85}{
\begin{tabular}{c}
$^{1}$Department of Agricultural and Food Sciences, University of Bologna, Bologna, Italy\\
$^{2}$Department of Architecture, Built Environment and Construction Engineering, Politecnico di Milano, Milan, Italy\\
$^{3}$Institute of Bio- and Geosciences, Plant Sciences, Forschungszentrum J\"ulich, J\"ulich, Germany\\
$^{4}$Institute for Advanced Simulation, Data Analytics and Machine Learning, Forschungszentrum J\"ulich, J\"ulich, Germany
\end{tabular}
}
}
\begin{document}
%\ninept
\ninept
\maketitle
\begin{abstract}

Accurate counting and localization of plants and their organs support phenotyping and yield estimation, yet target appearance, scale, and density vary widely across species and imaging conditions. Exemplar boxes specify the target without category-specific retraining, and point predictions identify the individual instances contributing to the count. We introduce AgriCountDINO, a parameter-efficient exemplar-guided framework for joint counting and localization. It conditions frozen multiscale DINOv3 features on exemplar appearance and size, then progressively decodes them into target points. Missed-object recovery extends supervision to targets overlooked by initial matching, and exemplar-adaptive point NMS filters duplicate predictions according to exemplar scale. With 8.4M trainable parameters, approximately one-tenth of TasselNetV4's, AgriCountDINO achieves a three-shot MAE of 11.92 on the TPC-268 benchmark, reducing counting error by 9.7\% while providing individual target locations. Trained only on TPC-268, it achieves a zero-shot MAE of 14.25 on unseen generic object categories in FSC-147, improving upon the best compared zero-shot method by 6.0\% without target-domain training or fine-tuning.
% 或者具体的数值 我们都可以改成百分比 ，这样更具有 论文的感觉，要不一堆数在这，他也看不懂
\end{abstract}
\begin{figure*}[t]
    \centering
    \includegraphics[width=0.90\textwidth]{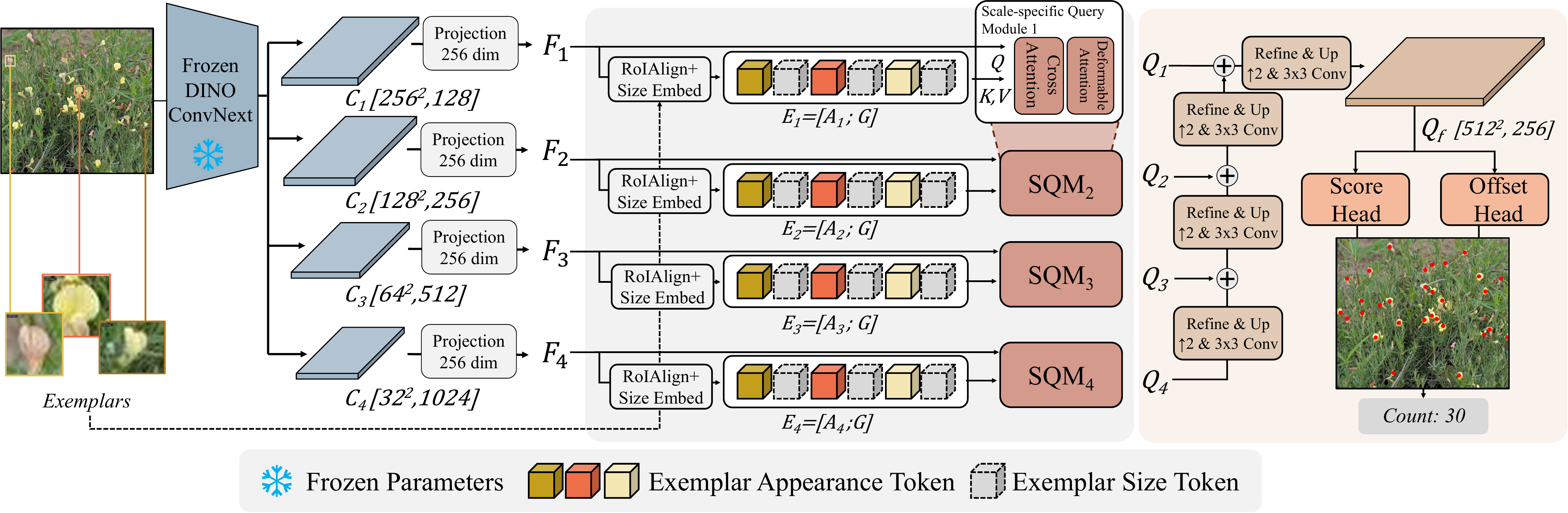}
    \caption{\textbf{Overview of AgriCountDINO.} Exemplar appearance and size tokens condition frozen multiscale features through scale-specific query modules (SQMs). Progressive coarse-to-fine fusion produces a high-resolution representation for point scoring and offset prediction. Feature dimensions are shown for a $1024\times1024$ input.}
    \label{fig:overview}
\end{figure*}

\begin{figure}[t]
    \centering
    \includegraphics[width=0.9\columnwidth]{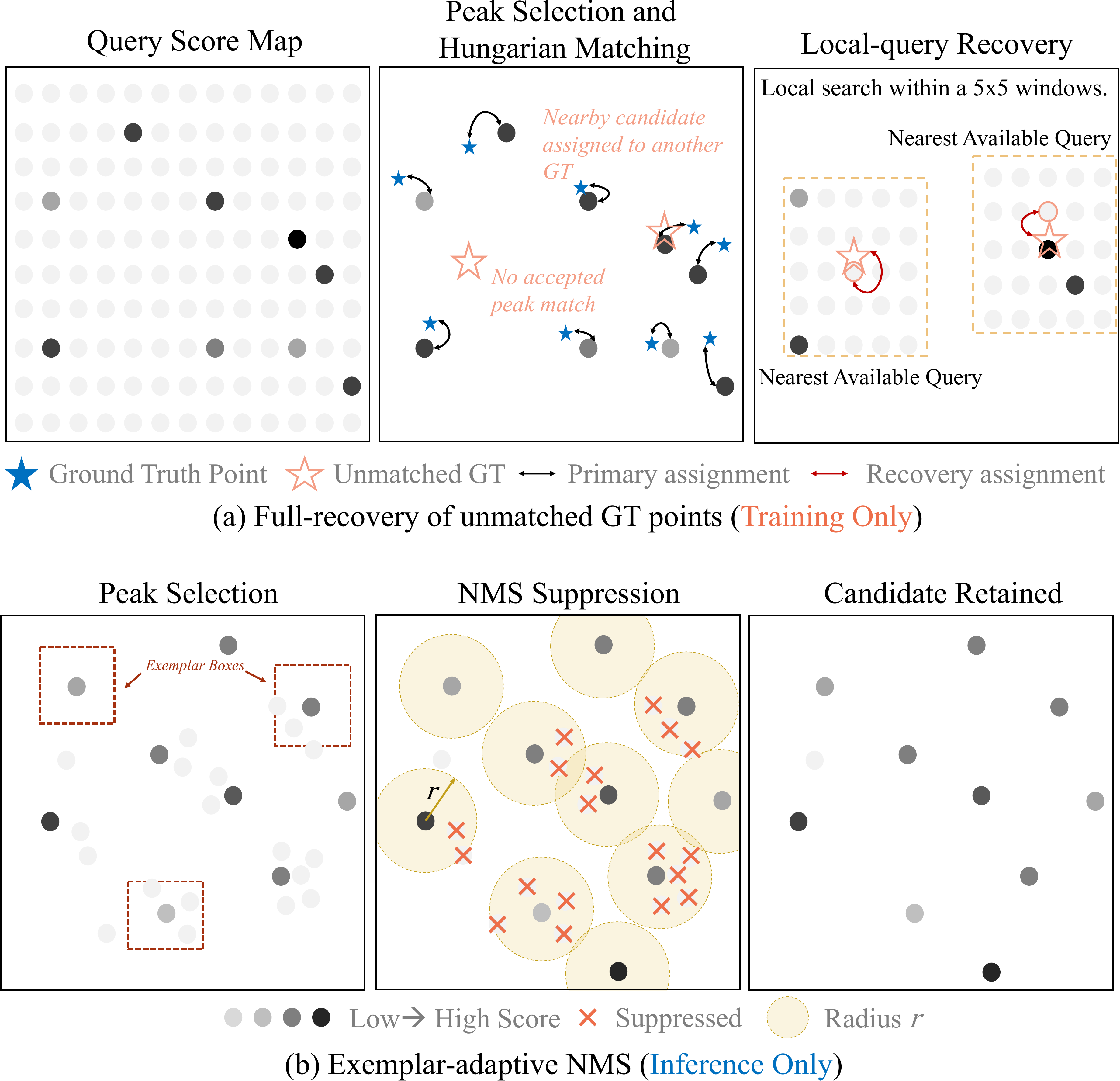}
    \caption{\textbf{Missed-object recovery and exemplar-adaptive point NMS.}
    (a) During training, unmatched targets are assigned to unique unused
    queries in their local neighborhoods. (b) During inference, nearby
    candidate points are suppressed using a radius determined by the
    exemplar scale.}
    \label{fig:recovery_nms}
\end{figure}

\section{Introduction}

Accurate counting and localization of plants and their organs are fundamental to agricultural image analysis. Count estimates support yield estimation and phenotyping, while instance locations enable spatial analysis and inspection of individual predictions. Agricultural targets vary substantially across species, organs, growth stages, scales, and imaging conditions, making it difficult to extend category-specific models to diverse targets~\cite{tasselnet4}. This diversity motivates exemplar-guided approaches, in which a few visual examples specify the objects of interest without requiring a separate model for each target category.

Recent counting methods can return individual target locations through point predictions, detections, or segmentation masks~\cite{song2021p2pnet,CountDETR,DAVE,GECO}. In agriculture, TasselNet introduced local count regression for maize tassels~\cite{lu2017tasselnet}, while TasselNetV4 combined local counting with exemplar guidance to address variation across scenes, scales, and species~\cite{tasselnet4}. TPC-268 provides exemplar boxes and point annotations for hundreds of fine-grained plant categories~\cite{TPC268}, allowing counting and localization to be evaluated together across diverse agricultural targets.

TasselNetV4's local-count predictions, however, do not explicitly identify individual targets. Exemplar-guided counting depends on relating the supplied examples to corresponding regions in the image~\cite{shi2022bmnet}. To also localize targets, these correspondences must retain the spatial detail needed to distinguish nearby instances as plant appearance and size vary. Self-supervised visual encoders provide transferable representations~\cite{IJEPA,DINO}; DINOv3, in particular, offers dense features that provide a basis for learning such correspondences~\cite{DINOv3}. This motivates \textbf{AgriCountDINO}\footnote{\url{https://github.com/Shengjie-Guo/AgriCountDINO}}, an exemplar-guided framework for joint plant counting and localization.

AgriCountDINO uses a frozen DINOv3-pretrained encoder~\cite{DINOv3,ConvNext} to extract multiscale image features. Exemplar appearance and size condition these features, which are fused from coarse to fine and decoded into instance points. Training updates only the exemplar-conditioning, fusion, and prediction modules. Missed-object recovery supervises annotated targets overlooked by initial peak matching. At inference, the exemplar scale determines which nearby point predictions are suppressed as duplicates. The retained points specify individual target locations, and their number gives the count. Counting and point-localization results on diverse plant categories, together with zero-shot transfer to unseen generic objects, suggest that the model captures transferable visual and semantic correspondences between exemplars and targets.

Our contributions are summarized as follows:
\begin{itemize}
    \item We introduce AgriCountDINO, an exemplar-guided model that counts and localizes diverse plants and plant organs with only 8.4M trainable parameters.

    \item We develop multiscale point prediction guided by exemplar appearance and size, with missed-object recovery during training and exemplar-adaptive duplicate suppression at inference.

    \item AgriCountDINO achieves a three-shot validation MAE of 11.92 on TPC-268, 9.7\% below TasselNetV4, and a zero-shot MAE of 14.25 on FSC-147, 6.0\% below the strongest compared zero-shot method.
\end{itemize}
\section{METHOD}
\label{sec:method}

Given an image $I$ and a set of exemplar boxes $\mathcal{E}$, AgriCountDINO predicts target locations $\hat{\mathcal{P}}$ and reports $|\hat{\mathcal{P}}|$ as the count. As shown in Fig.~\ref{fig:overview}, exemplar information is introduced at multiple levels of a frozen encoder before the conditioned features are aggregated for dense point prediction.

\subsection{Multiscale Exemplar Conditioning}
\label{sec:conditioning}

A frozen DINOv3-pretrained encoder~\cite{DINOv3,ConvNext} extracts four feature maps. For a $1024\times1024$ input, $C_1,C_2,C_3,C_4$ have sizes $256^2\times128$, $128^2\times256$, $64^2\times512$, and $32^2\times1024$, respectively. A separate $1\times1$ convolution projects each map to 256 channels, producing $\{F_l\}_{l=1}^{4}$.

At each scale, RoIAlign~\cite{he2017maskrcnn} extracts an appearance token from each of the $K$ exemplar boxes, while a shared lightweight MLP embeds each box's normalized width and height into a size token. The $K$ appearance tokens and $K$ size tokens form $E_l\in\mathbb{R}^{2K\times256}$. Inspired by previous work~\cite{GECO2}, a scale-specific query module (SQM) uses tokens from $F_l$ as queries and $E_l$ as keys and values. Cross-attention relates image locations to the exemplars' appearance and size cues, and single-scale deformable attention~\cite{zhu2021deformabledetr} refines the resulting features, yielding $\{Q_l\}_{l=1}^{4}$.
\subsection{Progressive Dense Point Decoding}
\label{sec:decoding}

The conditioned maps are fused from coarse to fine by upsampling the fused feature from the previous level and adding it to the next finer map. Let $\mathcal{R}_l$ denote $2\times$ bilinear upsampling followed by a $3\times3$ convolution and GELU. The decoder computes
\begin{equation}
\begin{aligned}
Z_4 &= Q_4,\\
Z_l &= Q_l+\mathcal{R}_l(Z_{l+1}),
       \quad l=3,2,1,\\
Q_f &= \mathcal{R}_f(Z_1).
\end{aligned}
\label{eq:fusion}
\end{equation}
Here, $Z_l$ is the fused feature at level $l$. For a $1024\times1024$ input, the final upsampling produces $Q_f$ with 256 channels at a resolution of $512\times512$.

Each position $i$ in $Q_f$ defines a feature vector $q_i$ and a reference point $r_i$ at the center of its grid cell in normalized image coordinates. Score and coordinate heads predict
\begin{equation}
s_i=h_s(q_i),\qquad
\hat{p}_i=\operatorname{clip}\!\left(
r_i+2\sigma\!\left(h_p(q_i)\right)-1,\,
0,\,1
\right).
\label{eq:point}
\end{equation}
Here, $h_s$ produces a target score, while $2\sigma(h_p(q_i))-1$ gives a two-dimensional displacement; $\operatorname{clip}$ bounds the resulting point to the image. A score peak may lie outside the grid cell containing its target center, particularly in crowded regions. Allowing the predicted point to cross cell boundaries lets that query localize the target center.
\subsection{Localization-Aware Learning and Inference} \label{sec:learning} 

The dense decoder provides a scored point at every position in $Q_f$. To establish unambiguous supervision, $3\times3$ local score maxima above the image-wise median are matched one-to-one with annotated points. The matching favors confident predictions close to target centers. For a selected candidate $k$ and annotation $g_j$, its cost is
\begin{equation}
\mathcal{C}_{kj}
=
\lambda_p\|\hat{p}_k-g_j\|_1
-\lambda_s\sigma(s_k).
\label{eq:matching}
\end{equation}
Here, $\hat{p}_k$ and $g_j$ are normalized image coordinates, $\sigma$ is the sigmoid function, and $\lambda_p$ and $\lambda_s$ balance localization and confidence. Matched pairs are retained only if their Euclidean distance in image pixels falls within an exemplar-dependent tolerance.

In crowded regions, closely spaced instances can produce overlapping responses on the score map, leaving fewer distinct peaks than annotated targets. Peak-based matching may therefore leave some targets without a positive query. Missed-object recovery (Fig.~\ref{fig:recovery_nms}(a)) extends supervision to these targets by assigning them unused dense queries. Candidate reference positions $r_i$ are searched within a $5\times5$ grid neighborhood of each unmatched annotation $g_j$, followed by a $7\times7$ search for any remaining targets. A joint one-to-one assignment minimizes the total reference-to-annotation $\ell_1$ distance while reserving queries accepted by the initial matching. This gives nearby targets distinct training queries.

The initial matching and missed-object recovery define the positive queries for training. Accepted matches and recovered queries receive positive score labels; the remaining selected peaks receive negative labels. Their supervision is organized as
\begin{equation}
\begin{aligned}
\mathcal{L}={}&
\mathcal{L}_{\mathrm{score}}
+\lambda_{\mathrm{loc}}\mathcal{L}_{\mathrm{loc}}\\
&+\rho_t\left(
\lambda_{\mathrm{xy}}\mathcal{L}_{\mathrm{rec}}^{\mathrm{xy}}
+\lambda_{\mathrm{peak}}\mathcal{L}_{\mathrm{rec}}^{\mathrm{peak}}
\right)
+\alpha(\mathcal{E})\mathcal{L}_{\mathrm{aux}}.
\end{aligned}
\label{eq:loss}
\end{equation}
The masked squared-error term $\mathcal{L}_{\mathrm{score}}$ supervises selected peaks and recovered queries. The $\ell_1$ term $\mathcal{L}_{\mathrm{loc}}$ trains point coordinates on accepted initial matches. For recovered queries, $\mathcal{L}_{\mathrm{rec}}^{\mathrm{xy}}$ penalizes coordinate error, while $\mathcal{L}_{\mathrm{rec}}^{\mathrm{peak}}$ encourages each query to score above nearby unassigned queries in its $3\times3$ neighborhood. Accepted and recovered positives are excluded from this comparison, and both recovery terms place greater weight on closely spaced targets relative to the exemplar scale. Recovered queries enter $\mathcal{L}_{\mathrm{score}}$ from the first epoch; $\rho_t$ gradually introduces only their coordinate and peak losses over the first eight epochs. The auxiliary loss $\mathcal{L}_{\mathrm{aux}}$ supervises predictions from $Q_1$ and is weighted by $\alpha(\mathcal{E})$ for small targets.

At inference, confident local score maxima yield candidate points, but several peaks may regress to the same target. Exemplar-adaptive point NMS (Fig.~\ref{fig:recovery_nms}(b)) processes candidates in descending score order and suppresses points near each retained prediction. Its radius is based on $\frac{1}{2}\min(\bar{w},\bar{h})$, where $\bar{w}$ and $\bar{h}$ are the mean exemplar width and height in image pixels. The retained points form $\hat{\mathcal{P}}$, whose cardinality gives the count.
\section{Experiments}

\begin{table*}[!t]
\centering
\caption{
\textbf{Comparison with representative state-of-the-art methods on TPC-268.}
Best and second-best results under each shot setting are shown in
\textbf{bold} and \underline{underline}, respectively.
}
\label{tab:tpc268_benchmark}

\setlength{\tabcolsep}{3.5pt}
\renewcommand{\arraystretch}{1.12}
\small

\begin{tabular}{lc|ccc|ccc|c}
\toprule
\multirow{2}{*}{Method}
& \multirow{2}{*}{Shot}
& \multicolumn{3}{c|}{Validation}
& \multicolumn{3}{c|}{Test}
& \multirow{2}{*}{Train. Params.$\downarrow$} \\
\cmidrule(lr){3-5}
\cmidrule(lr){6-8}
&
& MAE$\downarrow$
& RMSE$\downarrow$
& $R^2\uparrow$
& MAE$\downarrow$
& RMSE$\downarrow$
& $R^2\uparrow$
& \\
\midrule

% ==================== 3-shot ====================

LOCA (ICCV'23)\cite{LOCA}
& 3
& 17.26
& 53.19
& 0.75
& \textbf{17.51}
& \textbf{38.37}
& \textbf{0.78}
& \underline{11.37M} \\

CACViT (AAAI'24)\cite{CACVit}
& 3
& 16.63
& \underline{42.49}
& 0.82
& 22.04
& \underline{41.79}
& 0.73
& 99.77M \\

TasselNetV4 (ISPRS J.'26)\cite{tasselnet4}
& 3
& \underline{13.20}
& 43.93
& \underline{0.83}
& 22.95
& 51.36
& 0.67
& 84.67M \\

\rowcolor{blue!10}
\textbf{AgriCountDINO (Ours)}
& \textbf{3}
& \textbf{11.92}
& \textbf{28.55}
& \textbf{0.93}
& \underline{21.76}
& 49.58
& \underline{0.74}
& \textbf{8.40M} \\

\midrule

% ==================== 1-shot ====================

LOCA (ICCV'23)\cite{LOCA}
& 1
& 17.19$\pm$0.31
& 48.14$\pm$2.19
& 0.80$\pm$0.02
& \textbf{21.47$\pm$0.29}
& \textbf{42.36$\pm$0.72}
& \textbf{0.73$\pm$0.01}
& \underline{11.37M} \\

CACViT (AAAI'24)\cite{CACVit}
& 1
& 17.96$\pm$0.16
& 43.38$\pm$0.47
& 0.83$\pm$0.00
& 22.06$\pm$0.11
& \underline{42.97$\pm$0.81}
& \underline{0.71$\pm$0.01}
& 99.77M \\

TasselNetV4 (ISPRS J.'26)\cite{tasselnet4}
& 1
& \underline{13.49$\pm$0.02}
& \underline{41.30$\pm$0.46}
& \underline{0.85$\pm$0.00}
& 22.20$\pm$0.11
& 48.70$\pm$0.26
& 0.67$\pm$0.00
& 84.67M \\

\rowcolor{blue!10}
\textbf{AgriCountDINO (Ours)}
& \textbf{1}
& \textbf{13.16$\pm$0.07}
& \textbf{33.16$\pm$0.66}
& \textbf{0.90$\pm$0.00}
& \underline{22.03$\pm$0.18}
& 50.08$\pm$0.46
& 0.70$\pm$0.00
& \textbf{8.40M} \\

\bottomrule
\end{tabular}

\vspace{-2mm}
\end{table*}
\begin{table}[t]
\centering
\caption{\textbf{Spatial counting and localization on TPC-268 validation.}
G1--G3 denote GAME1--GAME3.}
\label{tab:tpc268_game_pr}

\resizebox{\columnwidth}{!}{%
\begin{tabular}{lcccccc}
\toprule
Method & Shot & G1$\downarrow$ & G2$\downarrow$ & G3$\downarrow$
& Prec.$\uparrow$ & Rec.$\uparrow$ \\
\midrule
TasselNetV4~\cite{tasselnet4}
& 3 & 17.71 & 22.75 & 33.33 & -- & -- \\
TasselNetV4~\cite{tasselnet4}
& 1 & 19.14 & 23.86 & 33.93 & -- & -- \\
\midrule
\rowcolor{blue!10}
\textbf{AgriCountDINO}
& 3 & \textbf{14.92} & \textbf{18.64} & \textbf{25.37}
& \textbf{.719} & \textbf{.729} \\
\rowcolor{blue!10}
\textbf{AgriCountDINO}
& 1 & 15.90 & 19.72 & 27.17
& .713 & .709 \\
\bottomrule
\end{tabular}%
}
\vspace{-2mm}
\end{table}
\begin{table}[t]
\centering
\caption{Three-shot in-domain and zero-shot TPC-268$\rightarrow$FSC-147 results.}
\label{tab:fsc_generalization}

\setlength{\tabcolsep}{4.5pt}
\renewcommand{\arraystretch}{1.08}
\small
\begin{tabular}{l|cc|cc}
\toprule
\multirow{2}{*}{Method}
& \multicolumn{2}{c|}{FSC-147$\rightarrow$FSC-147}
& \multicolumn{2}{c}{TPC-268$\rightarrow$FSC-147} \\
\cmidrule(lr){2-3}
\cmidrule(lr){4-5}
& MAE$\downarrow$ & RMSE$\downarrow$
& MAE$\downarrow$ & RMSE$\downarrow$ \\
\midrule

LOCA~\cite{LOCA}
& \underline{10.79} & \underline{56.97}
& \underline{15.16} & 109.15 \\

CACViT~\cite{CACVit}
& \textbf{9.13} & \textbf{48.96}
& 17.88 & \textbf{82.57} \\

TasselNetV4~\cite{tasselnet4}
& 16.16 & 110.56
& 19.17 & 116.24 \\

\midrule
\rowcolor{blue!10}
\textbf{AgriCountDINO}
& 11.62 & 85.07
& \textbf{14.25} & \underline{106.88} \\

\bottomrule
\end{tabular}

\vspace{-2mm}
\end{table}
\begin{table}[t]
\centering
\caption{Ablation study on the TPC-268 validation set under the 3-shot setting.}
\label{tab:ablation}
\setlength{\tabcolsep}{4pt}
\renewcommand{\arraystretch}{1.08}
\resizebox{\columnwidth}{!}{%
\begin{tabular}{lcccc}
\toprule
Configuration & MAE$\downarrow$ & G3$\downarrow$ & F1$\uparrow$ & Recall$\uparrow$ \\
\midrule
Supervised ConvNeXt-B prior
& 17.64 & 32.19 & 0.668 & 0.646 \\

w/o scale-wise exemplar conditioning
& 15.12 & 31.78 & 0.669 & 0.632 \\

w/o missed-object recovery
& 12.71 & 26.27 & 0.723 & 0.692 \\

\rowcolor{blue!10}
\textbf{Full model}
& \textbf{11.92} & \textbf{25.37} & \textbf{0.723} & \textbf{0.729}\\
\bottomrule
\end{tabular}
}
\end{table}
\begin{figure}[t]
    \centering
    \includegraphics[width=0.9\columnwidth]{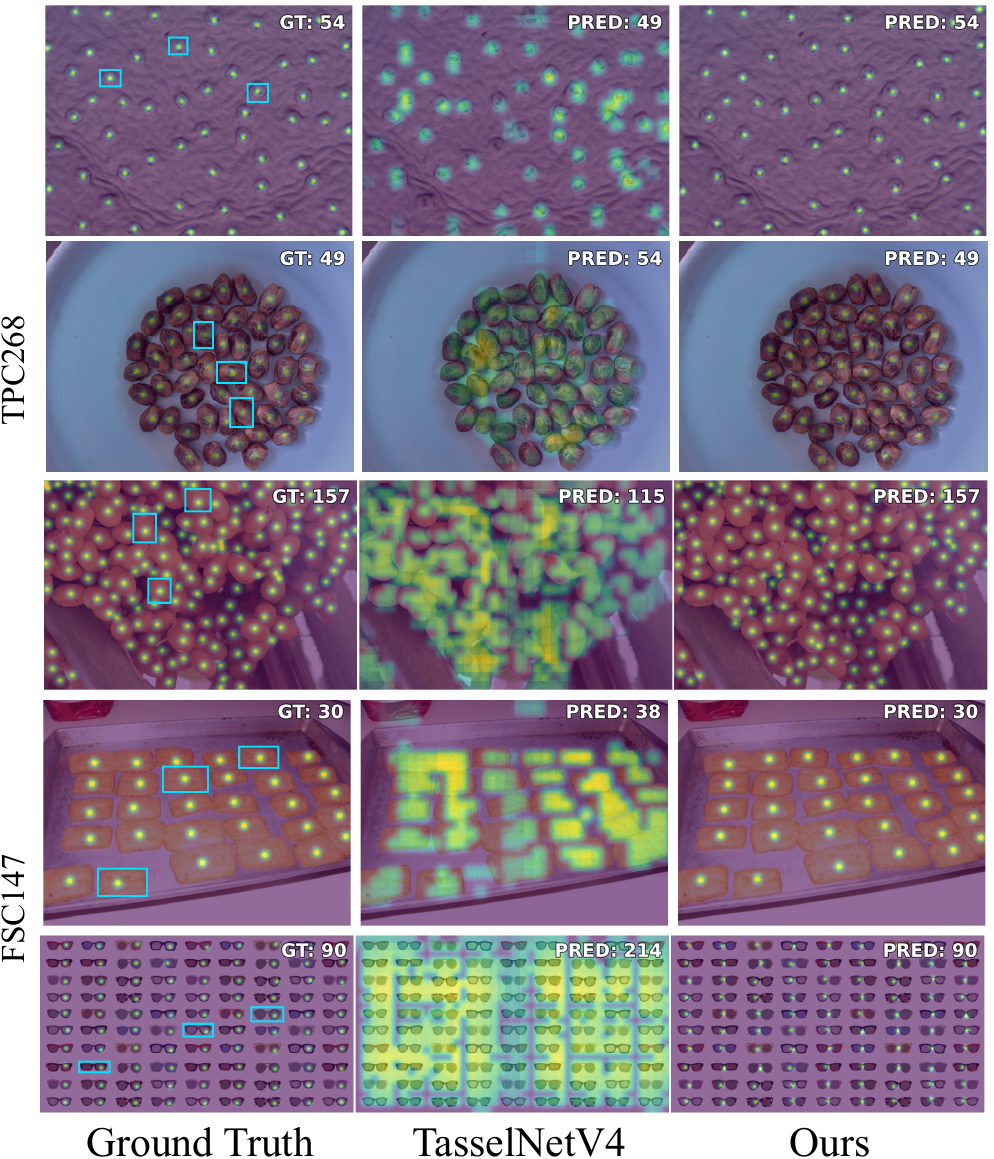}
    \caption{\textbf{Qualitative comparison with TasselNetV4.} Exemplar boxes are shown
    in the ground truth, and dots denote target locations. The top three rows
    are from TPC-268; the bottom two show zero-shot transfer to FSC-147.}
    \label{fig:instance}
\end{figure}

\subsection{Experimental Setup}
\subsection{Experimental Setup}

\noindent\textbf{Implementation details.}
AgriCountDINO uses a frozen DINOv3-pretrained ConvNeXt-B encoder~\cite{DINOv3,ConvNext}. Images are resized to $1024\times1024$, with up to three exemplar boxes used during training. The trainable parameters are optimized for 100 epochs with AdamW~\cite{adamw}, a learning rate and weight decay of $10^{-4}$, an effective batch size of 32, and BF16 mixed precision. The matching weights are $\lambda_p=1$ and $\lambda_s=2$; the loss weights are $\lambda_{\mathrm{loc}}=1$, $\lambda_{\mathrm{xy}}=0.5$, and $\lambda_{\mathrm{peak}}=0.05$. Recovery coordinate and peak losses are gradually introduced over the first eight epochs. The auxiliary weight is 0.3 when the smaller mean exemplar dimension is below 25 pixels, and zero otherwise.

\noindent\textbf{Datasets and metrics.}
AgriCountDINO is trained on the 7,000 images in the official TPC-268 training split and evaluated in one-shot and three-shot settings. MAE, RMSE, and $R^2$ measure image-level counting accuracy. To assess the spatial distribution of predicted counts, GAME-$L$~\cite{GAME} sums absolute count errors over $4^L$ image regions; we report levels $L=1,2,3$. This exposes regional errors that can cancel in the image-level count. Point-level precision and recall assess individual target localization. Zero-shot transfer is evaluated on the generic-object dataset FSC-147 without target-domain training or fine-tuning.
\begin{figure}[t]
    \centering
    \includegraphics[width=0.9
    \columnwidth]{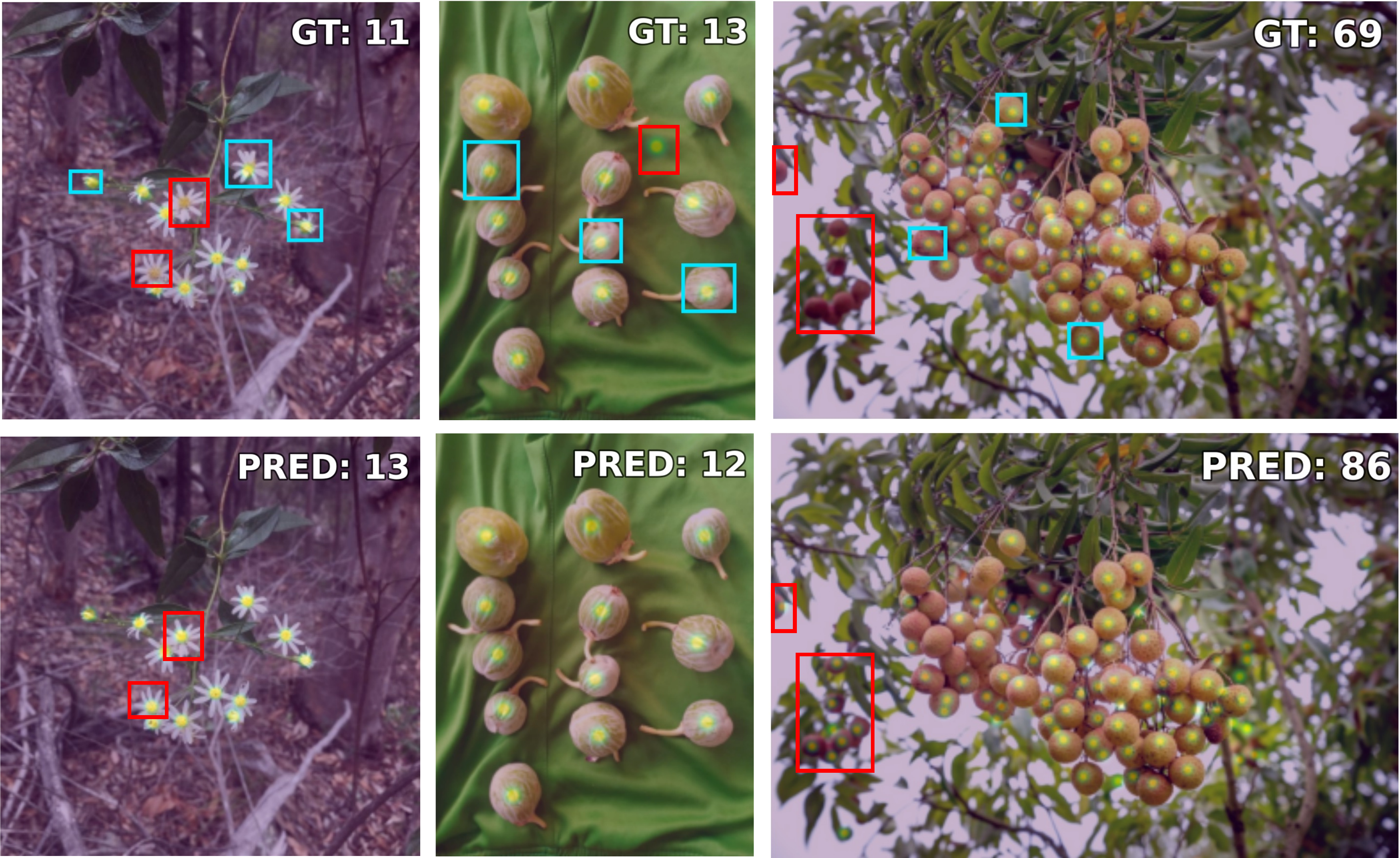}
    \caption{\textbf{Possible annotation discrepancies in TPC-268.} Top: annotations and exemplar boxes; bottom: AgriCountDINO predictions. Red boxes highlight apparent discrepancies}
    \label{fig:annotation_quality}

\end{figure}

\subsection{Counting Accuracy and Spatial Localization}
With 8.40M trainable parameters, approximately one tenth of TasselNetV4's, AgriCountDINO achieves the lowest TPC-268 validation MAE and RMSE in both exemplar settings (Table~\ref{tab:tpc268_benchmark}). In the three-shot setting, its MAE of 11.92 and RMSE of 28.55 reduce the corresponding errors of TasselNetV4 by 9.7\% and 35.0\%. With one exemplar, it obtains a validation MAE of 13.16. On the test split, the three-shot and one-shot MAEs are 21.76 and 22.03, compared with 22.95 and 22.20 for TasselNetV4.

To examine whether accurate counts are accompanied by accurate target placement, we also evaluate predicted point localization. In the three-shot setting, AgriCountDINO reduces GAME3 from 33.33 for TasselNetV4 to 25.37, a 23.9\% reduction at the finest reported grid level (Table~\ref{tab:tpc268_game_pr}). Its predicted points achieve 0.719 precision and 0.729 recall. The examples in Fig.~\ref{fig:instance} show points aligned with individual targets, including closely spaced instances where TasselNetV4 produces broader counting responses.

Remaining errors on the validation and test sets may arise from both difficult target configurations and annotation quality. Dense or visually ambiguous instances can challenge point prediction, while Fig.~\ref{fig:annotation_quality} shows examples of apparent discrepancies between visible targets and point annotations. A prediction on a plausible but unannotated target may therefore increase the measured error. 

\subsection{Zero-Shot Cross-Dataset Transfer}

Trained only on TPC-268, AgriCountDINO transfers directly to unseen generic object categories in FSC-147 without target-domain training or fine-tuning. Table~\ref{tab:fsc_generalization} reports a zero-shot MAE of 14.25, improving on LOCA's 15.16 by 6.0\%. The FSC-147 examples in Fig.~\ref{fig:instance} further show that the model localizes individual targets in these unseen categories. These results suggest that the model captures transferable semantic correspondences between exemplars and target objects, enabling counting and localization beyond the agricultural categories seen during training.

\subsection{Ablation Study}

The ablations in Table~\ref{tab:ablation} reveal distinct roles for the three components. Replacing DINOv3 with supervised ConvNeXt-B causes the largest increase in MAE (11.92 to 17.64), suggesting that the DINOv3 representation provides a stronger basis for relating exemplars to diverse plant targets. Removing scale-wise conditioning worsens both MAE and GAME3, indicating that exemplar guidance across the feature hierarchy contributes to accurate global and regional counts. Without missed-object recovery, point recall falls from 0.729 to 0.692 and MAE increases, consistent with more targets being omitted after initial peak matching. F1 remains unchanged, so the clearest contribution of recovery is improved instance coverage.

\section{Conclusions}

We introduced AgriCountDINO, a parameter-efficient model for exemplar-guided counting and localization of plants and plant organs. It conditions frozen multiscale DINOv3 features on exemplar appearance and size, then predicts individual target points whose number gives the count. With 8.4M trainable parameters, AgriCountDINO achieves lower validation counting and spatial errors than TasselNetV4 on TPC-268. It also obtains the lowest zero-shot MAE among the compared methods on FSC-147 without target-domain fine-tuning. These results demonstrate AgriCountDINO's ability to combine accurate counting, instance localization, and transfer beyond the agricultural categories used for training.

\section{Acknowledgment}
This work was co-funded by the European Union through the FutureData4EU project (Grant Agreement No. 101126733). The views expressed are those of the authors and do not necessarily reflect those of the European Union or REA. We acknowledge ISCRA for awarding access to the LEONARDO supercomputer, owned by the EuroHPC Joint Undertaking and hosted by CINECA (Italy).
\bibliographystyle{IEEEbib}
\bibliography{refs}
\end{document}